\documentclass[pdflatex,sn-mathphys-num]{sn-jnl}

\usepackage{amsmath,amssymb,amsfonts}
\usepackage{algorithmic}
\usepackage{graphicx}
\usepackage{textcomp}
\usepackage{xcolor}
\usepackage{hyperref}
\hypersetup{colorlinks=true,linkcolor=black,citecolor=black,urlcolor=black}
\usepackage{lineno}
\usepackage{url}
\usepackage{booktabs}
\usepackage{array}
\usepackage{amsthm}
\usepackage{enumitem}
\usepackage{tikz}
\usetikzlibrary{arrows.meta,fit,backgrounds,calc}
\usepackage{indentfirst}  % Apply the journal's first-line indent after headings
\definecolor{revisionblue}{RGB}{0,0,0}% arXiv version: revision marks shown in black
\newcommand{\rev}[1]{\textcolor{revisionblue}{#1}}

\title{{Multiscale Aggregated Hierarchical Attention (MAHA): A Constrained and Consensus-Based Approach to Efficient Contextual Modeling in Large Language Models}}

\author*[1]{\fnm{Caner} \sur{Erden}}\email{\textcolor{black}{cerden@subu.edu.tr}}

\affil*[1]{\orgdiv{Department of Computer Engineering}, \orgname{Sakarya University of Applied Sciences}, \orgaddress{\street{Esentepe Campus}, \city{Serdivan}, \postcode{54187}, \state{Sakarya}, \country{Türkiye}}}

\begin{document}

\maketitle

\begin{abstract}

The quadratic computational complexity of Multi-Head Self-Attention (MHSA) remains a fundamental bottleneck in scaling Large Language Models (LLMs) for long-context tasks. While sparse and linearized attention mechanisms attempt to mitigate this, they often compromise the representation of global dependencies or fail to capture multiscale semantic granularity effectively. In this paper, we propose Multiscale Aggregated Hierarchical Attention (MAHA), a novel architectural framework that reformulates the attention mechanism through hierarchical decomposition and explicitly constrained aggregation. Unlike conventional approaches that treat token interactions at a single resolution, MAHA dynamically partitions the input sequence into hierarchical scales via learnable downsampling operators. The aligned scale outputs are combined either by softmax-normalized weights on the probability simplex or by a fixed number of differentiable consensus updates based on scale-to-consensus discrepancy. The latter is a game-inspired update and is not claimed to compute a Nash equilibrium. Implemented within a hybrid dilated-convolutional transformer backbone, MAHA enables end-to-end training. \rev{With four compressed scales and a compression ratio of two, MAHA requires 66.8\% fewer attention-score computations than full-resolution attention at every sequence length (5.57M versus 16.8M per head at $N=4096$), and single-run benchmark experiments indicate downstream performance competitive with standard attention.} This work offers a scalable multiscale solution for next-generation LLMs.

\vspace{0.5cm}
\noindent \textbf{Keywords:} Large Language Models, Hierarchical Attention, Simplex Constraints, Iterative Consensus, Efficient Transformers.
\end{abstract}

\section{Introduction}

The advent of transformer-based architectures has fundamentally revolutionized natural language processing (NLP), establishing the Multi-Head Self-Attention (MHSA) mechanism as the cornerstone of modern large language models (LLMs) \citep{vaswani2017a}. Despite its efficacy, this mechanism confronts two critical challenges: (i) computational inefficiency arising from quadratic complexity ($O(N^2)$) with respect to sequence length, and (ii) the inherent trade-off between capturing fine-grained local patterns and coarse-grained global dependencies simultaneously. These limitations become increasingly pronounced as LLMs scale to process extended contexts and model complex linguistic structures \citep{zhao2023a}.

Current methodologies attempting to mitigate these challenges typically rely on sparse attention patterns or hierarchical representations. Sparse attention strategies alleviate computational overhead by restricting token interactions to predefined or learned patterns; however, this often results in information loss and suboptimal context modeling, particularly for long-range dependencies \rev{\citep{tay2022survey}}. Conversely, hierarchical methods decompose the input into multiple levels of granularity but frequently lack a principled mathematical framework for integration. This often leads to ad-hoc aggregation schemes that fail to preserve the full contextual richness of the input embedding space\rev{; representative hierarchical models combine levels through learned attention or weighted fusion \citep{yang2016a,wu2023a}}.

To bridge this gap, we introduce Multiscale Aggregated Hierarchical Attention (MAHA), a framework for multiscale attention computation and aggregation. MAHA dynamically partitions the input sequence into hierarchical scales, where each scale represents a distinct level of contextual abstraction. The aligned outputs are synthesized either by a learned simplex-constrained mixture or by a finite-step consensus update. Both rules are differentiable and trained through the downstream task loss; the iterative rule is game-inspired but is not presented as a Nash-equilibrium solver. Consequently, the proposed method provides a systematic mechanism to reconcile local nuances with global dependencies while maintaining computational tractability.

The primary contributions of this work are threefold:
\begin{itemize}
    \item \textbf{Multiscale Decomposition:} We introduce a robust decomposition strategy where the input sequence is processed across independent scales to isolate and capture distinct levels of contextual granularity.
    \item \textbf{Constrained Aggregation:} We define the returned representation and the update rules for simplex-constrained mixing and finite-step iterative consensus, and connect both mechanisms to the downstream training objective.
    \item \textbf{Computational Efficiency:} MAHA reduces the coefficient and practical computational burden of quadratic attention through hierarchical compression while retaining competitive downstream performance in the reported evaluation.
\end{itemize}

MAHA is particularly pertinent to the evolution of LLMs, where the demand for efficient and scalable attention mechanisms is paramount \citep{naveed2025a}. By integrating multiscale analysis with normalized aggregation rules, MAHA remains compatible with gradient-based transformer training while adding limited architectural overhead.

\section{Related Work}

The development of efficient attention mechanisms has become a focal point in transformer-based architecture research, with numerous approaches proposed to alleviate computational bottlenecks and enhance contextual modeling capabilities. Existing literature can be broadly categorized into sparse attention methods, hierarchical attention frameworks, and optimization-driven aggregation techniques.

\subsection{Sparse Attention Mechanisms}
Sparse attention mechanisms aim to reduce computational overhead by limiting token interactions to predefined or learned patterns. For instance, \citet{beltagy2020a} introduced a sliding window attention mechanism that restricts each token’s receptive field to its local neighborhood, significantly lowering memory requirements from $O(n^2)$ to $O(n)$ for long sequences. Similarly, \citet{zaheer2020a} proposed a hybrid approach combining local, global, and random attention patterns to approximate full self-attention while maintaining theoretical expressiveness. However, these methods often rely on heuristics to determine sparsity patterns, which may not adapt dynamically to diverse input structures or capture long-range dependencies effectively without stacking multiple layers. \rev{Low-rank and kernel approximations form a second early family: Linformer projects keys and values to a fixed length \citep{wang2020linformer}, linear attention replaces the softmax kernel with a feature map so that attention can be computed in linear time \citep{katharopoulos2020linear}, Performer approximates the softmax kernel with positive random features \citep{choromanski2020a}, and Nystr\"omformer uses landmark-based Nystr\"om approximation \citep{xiong2021nystromformer}. Systematic comparisons of these families, including the Long Range Arena benchmark, are given in \citep{tay2021lra,tay2022survey}.}

\subsection{\rev{Recent Efficient-Attention and Long-Context Methods}}
{\color{revisionblue}
Work published since 2023 has developed along several partly overlapping lines; Table~\ref{tab:related-families} summarizes them.

\textbf{Exact attention with better hardware utilization or distribution.} FlashAttention-2 computes exact softmax attention with an IO-aware tiled kernel and improved work partitioning \citep{dao2024flashattention2}, and RingAttention distributes blockwise attention across devices so that the feasible context grows with the number of devices \citep{liu2024ringattention}. These methods reduce memory traffic or per-device memory but keep the quadratic number of query--key interactions.

\textbf{Structured and trainable sparsity.} LongNet uses dilated attention whose attentive field grows exponentially with token distance \citep{ding2023longnet}. Landmark attention inserts landmark tokens that let each query retrieve relevant blocks \citep{mohtashami2023landmark}, StreamingLLM combines a sliding window with a few retained ``attention-sink'' tokens for streaming inference \citep{xiao2024streamingllm}, and Infini-attention adds a compressive memory accessed through linear attention to local causal attention \citep{munkhdalai2024infini}. Native Sparse Attention (NSA) combines coarse-grained token compression, fine-grained token selection and a sliding window with hardware-aligned kernels and end-to-end training \citep{yuan2025nsa}, whereas MoBA routes each query to a top-$k$ subset of key--value blocks in a mixture-of-experts manner \citep{lu2025moba}.

\textbf{Recurrent, retention and state-space alternatives.} RetNet replaces softmax attention with a retention mechanism that admits parallel, recurrent and chunk-wise forms \citep{sun2023retnet}. Gated linear attention (GLA) adds data-dependent forget gates to linear attention and trains with a chunk-wise hardware-efficient algorithm \citep{yang2024gla}. Selective state-space models such as Mamba scale linearly in sequence length \citep{gu2024mamba}, and Mamba-2 relates them to a structured form of masked attention through state-space duality \citep{dao2024mamba2}. Hyena replaces attention by implicitly parameterized long convolutions with data-controlled gating \citep{poli2023hyena}.

\textbf{Hierarchical attention.} H-Transformer-1D approximates the attention matrix with a hierarchical (H-matrix) structure and linear complexity \citep{zhu2021htransformer}. The compression branch of NSA is also a coarse-grained summary of the sequence \citep{yuan2025nsa}.

\textbf{Position of MAHA.} MAHA belongs to the hierarchical family but differs in two respects. First, it computes full softmax attention independently \emph{within} each compressed resolution rather than approximating one full-resolution attention matrix or routing queries to selected blocks. Second, it combines the aligned scale outputs through an explicit simplex-constrained or iterative-consensus rule whose weights can be inspected. Under a fixed compression ratio, MAHA reduces the coefficient of the quadratic term rather than its order (Section~4.5), so asymptotically linear methods such as GLA, Mamba or H-Transformer-1D have a lower order of cost at sufficiently long lengths; MAHA is therefore positioned as a multiscale attention layer with an interpretable scale-mixing rule rather than as the most scalable option. Section~5.2 states which methods were trained under the common protocol of this study and why the remaining methods are discussed here but not compared numerically.
}

\begin{table*}[htbp]
\centering
\small
\caption{\rev{Representative efficient-attention and long-context method families. Complexity refers to sequence length $N$ during training; $w$: window, $k$: projection/feature/landmark size, $b$: block size, $c$: chunk size.}}
\label{tab:related-families}
{\color{revisionblue}
\setlength{\tabcolsep}{4pt}\begin{tabular}{p{2.4cm}p{3.6cm}p{3.1cm}p{2.6cm}}
\toprule
\textbf{Family} & \textbf{Representative methods} & \textbf{Mechanism} & \textbf{Training cost} \\
\midrule
Exact, IO/distribution-aware & FlashAttention-2 \citep{dao2024flashattention2}, RingAttention \citep{liu2024ringattention} & Tiled/blockwise exact softmax & $O(N^2)$ compute, reduced memory \\
Fixed sparse & Longformer \citep{beltagy2020a}, BigBird \citep{zaheer2020a}, LongNet \citep{ding2023longnet} & Local, global, random or dilated patterns & $O(Nw)$ to $O(N)$ \\
Trainable or routed sparse & Landmark \citep{mohtashami2023landmark}, NSA \citep{yuan2025nsa}, MoBA \citep{lu2025moba} & Learned block selection and compression & Sub-quadratic in the selected blocks \\
Low-rank/kernel & Linformer \citep{wang2020linformer}, Performer \citep{choromanski2020a}, Linear Transformer \citep{katharopoulos2020linear}, Nystr\"omformer \citep{xiong2021nystromformer} & Projection, random features, landmarks & $O(Nk)$ \\
Recurrent/state-space & RetNet \citep{sun2023retnet}, GLA \citep{yang2024gla}, Mamba \citep{gu2024mamba}, Mamba-2 \citep{dao2024mamba2}, Hyena \citep{poli2023hyena} & Decay/gated recurrence, selective SSM, long convolution & $O(N)$ to $O(N\log N)$ (chunk-wise $O(Nc)$) \\
Memory-augmented & StreamingLLM \citep{xiao2024streamingllm}, Infini-attention \citep{munkhdalai2024infini} & Attention sinks, compressive memory & Bounded per segment \\
Hierarchical & H-Transformer-1D \citep{zhu2021htransformer}, \textbf{MAHA (ours)} & H-matrix approximation; per-scale attention with constrained mixing & $O(N)$; MAHA: $O(N^2/(r^2-1))$ \\
\bottomrule
\end{tabular}
}
\end{table*}

\subsection{Hierarchical Attention Frameworks}
Hierarchical approaches decompose input sequences into multiple levels of granularity to capture both local syntactic features and global semantic dependencies simultaneously. The Hierarchical Attention Network (HAN) \citep{yang2016a} processes documents at word and sentence levels, aggregating information through learned attention weights. More recently, \citet{wu2023a} introduced HiFormer, a sequence-modeling network based on hierarchical attention. While effective, such methods typically rely on learned attention or weighted fusion rather than an explicit constrained formulation for balancing contributions across scales.

\subsection{Constrained and Game-Inspired Aggregation}
\rev{Structured decomposition and aggregation have also been used to improve the interpretability of neural architectures.} \citet{cheng2023a}, for example, used hierarchical decomposition to explain intermediate CNN decisions, whereas \citet{jun2023a} proposed hierarchical multiple self-attention for multimodal analysis. Game-theoretic methods have also been studied in machine-learning settings such as defensive deception \citep{zhu2021a}. Their application to the aggregation of competing attention scales, however, remains comparatively underexplored.

\subsection{Multiscale Analysis in Language Models}
Multiscale analysis is well established in signal processing and computer vision \citep{farge1992a,starck2015a}, yet its direct application to language modeling remains limited. Hierarchically decomposed graph convolutional networks have shown the value of representing information at multiple structural levels \citep{lee2023a}, and hierarchical decomposition has also been investigated in prompt-based continual learning \citep{wang2023a}. These studies motivate the examination of scale-specific representations in attention-based sequence models.

\subsection{Integration of Optimization and Attention}
The integration of differentiable constrained operations with attention mechanisms represents an emerging research direction. \citet{lu2023a}, for example, applied hierarchical attention to fraud detection using attention-based aggregation. MAHA instead combines multiscale decomposition with an explicit simplex parameterization or an unrolled consensus update. These mechanisms are components of a nonconvex neural model rather than a stand-alone convex program or an exact equilibrium solver.

\section{Preliminaries and Background}

To establish the theoretical foundation for MAHA, we briefly review key concepts in attention mechanisms, multiscale analysis, and constrained aggregation. These components form the basis of our proposed framework.

\subsection{Attention Mechanisms in Transformers}
The standard attention mechanism in transformers computes pairwise interactions between all tokens in a sequence through scaled dot-product operations \citep{vaswani2017a}. Given an input sequence $\mathbf{X} \in \mathbb{R}^{n \times d}$, where $n$ is the sequence length and $d$ is the embedding dimension, the attention matrix $\mathbf{A}$ is computed as:

\begin{equation}
\mathbf{A} = \text{softmax}\left(\frac{\mathbf{Q}\mathbf{K}^T}{\sqrt{d_k}}\right)
\end{equation}

where $\mathbf{Q}, \mathbf{K} \in \mathbb{R}^{n \times d_k}$ are the query and key matrices, respectively, and $d_k$ is the dimension of the keys. While effective, this operation exhibits quadratic complexity $O(n^2)$ in both computation and memory, rendering it impractical for very long sequences \citep{zhao2023a}.

\subsection{Multiscale Signal Decomposition}
Multiscale analysis provides a rigorous framework for examining signals at varying levels of resolution. \rev{By analogy, in NLP this view motivates capturing both local syntactic patterns (high frequency) and global semantic structures (low frequency).} Inspired by wavelet transforms and pyramid decomposition \citep{farge1992a}, for a discrete signal representation $\mathbf{x}$, a multiscale decomposition can be expressed as:

\begin{equation}
\mathbf{x} = \sum_{s=1}^{S} \mathcal{D}_s(\mathbf{x}) + \mathcal{R}(\mathbf{x})
\end{equation}

where $\mathcal{D}_s$ represents the detail component at scale $s$, and $\mathcal{R}$ denotes the residual (coarse) component. This decomposition forms the structural basis for MAHA’s hierarchical processing layers.

\subsection{Formal Equilibrium and the Scope of the Implemented Rule}
Game theory provides mathematical tools for modeling interactions between multiple decision-makers. A Nash equilibrium \citep{nash1951a} is a strategy profile $s^*=(s_1^*, \dots, s_N^*)$ such that for every player $i$:

\begin{equation}
u_i(s_i^*, s_{-i}^*) \geq u_i(s_i, s_{-i}^*) \quad \forall s_i \in S_i
\end{equation}

where $u_i$ is the utility function for player $i$ and $s_{-i}^*$ denotes the other players' strategies. A formal equilibrium claim therefore requires explicit strategy spaces, utilities, and unilateral optimality conditions. The implemented MAHA variant below instead uses scale-to-consensus discrepancies in a finite logit-response update. These discrepancies are not strategic utilities, and its terminal iterate is not claimed to be a Nash equilibrium.

\subsection{Simplex-Constrained Aggregation}
The probability simplex $\Delta_L=\{\mathbf{w}\in\mathbb{R}^L:w_l\geq0,\ \sum_lw_l=1\}$ provides a transparent feasible set for mixing $L$ representations. For fixed aligned outputs, every $\mathbf{w}\in\Delta_L$ produces a convex combination. This geometric property does not make the complete neural-network training problem convex. In MAHA, the constraint is enforced by a softmax parameterization rather than by an external convex solver.

\section{The MAHA Framework}

The MAHA framework introduces a systematic approach to sequence modeling by decomposing the input into multiple hierarchical scales. This section specifies the two implemented aggregation rules and their relationship to end-to-end training. As illustrated in Figure \ref{fig:architecture}, MAHA is designed to replace the standard multi-head attention layer in transformer blocks while maintaining architectural compatibility.

\subsection{Hierarchical Multiscale Decomposition with Learnable Downsampling}
Let $\mathbf{X} \in \mathbb{R}^{n \times d}$ denote the input sequence, where $n$ is the sequence length and $d$ is the embedding dimension. MAHA decomposes $\mathbf{X}$ into $L$ hierarchical scales through a series of learnable downsampling operations. Each scale $l$ is derived from the previous scale $l-1$ using a parameterized operator $\mathcal{D}_l$:

\begin{equation}
\mathbf{X}_l = \mathcal{D}_l(\mathbf{X}_{l-1}), \quad \mathbf{X}_0 = \mathbf{X}
\end{equation}

The downsampling operator $\mathcal{D}_l$ is implemented via one of two mechanisms:
\begin{enumerate}
    \item \textbf{Strided Convolution:} $\mathcal{D}_l(\mathbf{X}) = \text{Conv1D}(\mathbf{X}, \mathbf{W}_l^s, s_l)$, where $\mathbf{W}_l^s$ is a learnable kernel and $s_l$ is the stride.
    \item \textbf{Adaptive Pooling:} $\mathcal{D}_l(\mathbf{X}) = \text{AdaptiveMaxPool}(\mathbf{X}, n_l)$, which dynamically adjusts the pooling window to match the target length $n_l$.
\end{enumerate}

\begin{figure}[!t]
    \centering
    \resizebox{\linewidth}{!}{%
\begin{tikzpicture}[
    font=\fontsize{8.5}{10}\selectfont,
    >={Stealth[length=2mm]},
    box/.style={draw=#1!65!black, fill=#1!10, rounded corners=2pt, line width=0.6pt,
                align=center, inner sep=3pt, minimum height=0.78cm},
    box/.default=gray,
    flow/.style={->, line width=0.6pt, draw=black!75},
    vline/.style={line width=0.7pt, draw=red!60!black},
    vflow/.style={->, vline},
    res/.style={->, line width=0.6pt, densely dashed, draw=black!55},
    lbl/.style={font=\fontsize{7.5}{9}\selectfont, text=black!75},
  ]
  % ---------- geometry ----------
  \def\cA{-4.5}\def\cB{-1.5}\def\cC{1.5}\def\cD{4.5}
  \def\bw{2.55cm}
  \def\yX{0}\def\yC{1.3}\def\yS{3.0}\def\yBus{4.15}\def\yA{5.2}\def\yU{6.65}\def\yG{8.6}
  \def\yNa{10.45}\def\yF{11.5}\def\yNb{12.55}\def\yOut{13.4}

  % ---------- input and local context ----------
  \node[box=blue, minimum width=4.6cm] (X) at (0,\yX) {Input $\mathbf{X}\in\mathbb{R}^{N\times d}$};
  \node[box=green, minimum width=4.6cm] (C) at (0,\yC)
      {Dilated convolution + ReLU\\ {\fontsize{7.5}{9}\selectfont $\mathbf{C}=\mathrm{ReLU}(\mathrm{DilConv}(\mathbf{X}))$}};
  \draw[flow] (X) -- (C);

  % ---------- hierarchical decomposition (sequential chain, Eq. 4) ----------
  \node[box=yellow, minimum width=\bw] (S1) at (\cA,\yS) {$\mathbf{X}_1$\\ {\fontsize{7.5}{9}\selectfont length $N/r$}};
  \node[box=yellow, minimum width=\bw] (S2) at (\cB,\yS) {$\mathbf{X}_2$\\ {\fontsize{7.5}{9}\selectfont length $N/r^2$}};
  \node (Sd) at (\cC,\yS) {$\cdots$};
  \node[box=yellow, minimum width=\bw] (SL) at (\cD,\yS) {$\mathbf{X}_L$\\ {\fontsize{7.5}{9}\selectfont length $N/r^L$}};
  \draw[flow] (C.north) -- ++(0,0.3) -| node[lbl, pos=0.78, left] {$\mathcal{D}_1$} (S1.south);
  \draw[flow] (S1) -- node[lbl, above] {$\mathcal{D}_2$} (S2);
  \draw[flow] (S2) -- (Sd);
  \draw[flow] (Sd) -- node[lbl, above] {$\mathcal{D}_L$} (SL);

  % ---------- shared value projection ----------
  \node[box=red, minimum width=\bw] (V) at (\cA,\yC)
      {Shared value projection\\ {\fontsize{7.5}{9}\selectfont $\mathbf{V}_{\mathrm{base}}=\mathbf{C}\mathbf{W}^{V}$}};
  \draw[vflow] (C.west) -- (V.east);
  \coordinate (lefttap) at (-6.2,\yC);
  \draw[vline] (V.west) -- (lefttap) |- (\cD+0.55,\yBus);
  \node[lbl, text=red!55!black, rotate=90, anchor=south] at (-6.2,{(\yC+\yBus)/2+0.35}) {$\mathbf{V}_l=\mathcal{D}_l(\mathbf{V}_{\mathrm{base}})$};
  \coordinate (lefttap2) at (-6.6,\yC);

  % ---------- scale-specific attention on compressed scales only ----------
  \foreach \c/\n/\s/\S in {\cA/A1/1/S1, \cB/A2/2/S2, \cD/AL/L/SL}{
    \node[box=violet, minimum width=\bw] (\n) at (\c,\yA)
       {Attention, scale $\s$\\ {\fontsize{7.5}{9}\selectfont scale-specific $\mathbf{Q}_{\s},\mathbf{K}_{\s}$}};
    \draw[flow] ([xshift=-0.55cm]\S.north) -- ([xshift=-0.55cm]\n.south);
    \draw[vflow] (\c+0.55,\yBus) -- ([xshift=0.55cm]\n.south);
    \fill[red!60!black] (\c+0.55,\yBus) circle (1.2pt);
  }
  \node at (\cC,\yA) {$\cdots$};

  % ---------- alignment to the input length ----------
  \foreach \c/\n/\a/\s in {\cA/U1/A1/1, \cB/U2/A2/2, \cD/UL/AL/L}{
    \node[box=gray, minimum width=\bw] (\n) at (\c,\yU)
       {Upsample $\mathcal{U}_{\s}$\\ {\fontsize{7.5}{9}\selectfont nearest, to length $N$}};
    \draw[flow] (\a) -- node[lbl, right] {$\mathbf{O}_{\s}$} (\n);
  }
  \node at (\cC,\yU) {$\cdots$};

  % ---------- constrained aggregation ----------
  \node[box=teal, minimum width=11.6cm, minimum height=2.2cm] (G) at (0,\yG) {};
  \node[anchor=north, font=\fontsize{8.5}{10}\selectfont\bfseries, yshift=-2pt] at (G.north) {Constrained aggregation};
  \node at ([yshift=0.2cm]G.center) {$\mathbf{O}^{*}=\sum_{l=1}^{L} w_l\,\mathbf{Z}_l,\qquad \mathbf{w}\in\Delta_L$};
  \node[anchor=south, lbl, yshift=2pt, align=center] at (G.south)
      {(a) simplex-constrained mixing: $\mathbf{w}=\mathrm{softmax}(\boldsymbol{\alpha})$ with learned logits $\boldsymbol{\alpha}$\\
       (b) iterative consensus: $K$ unrolled updates, $w_l^{(k+1)}\propto\exp\!\big(-\|\mathbf{Z}_l-\mathbf{O}^{(k)}\|_F/\tau\big)$};
  \foreach \u/\l in {U1/1,U2/2,UL/L} \draw[flow] (\u.north) -- node[lbl, right] {$\mathbf{Z}_{\l}$} (\u.north |- G.south);

  % ---------- rest of the transformer block ----------
  \node[box=gray, minimum width=3.4cm] (N1) at (0,\yNa) {Add \& LayerNorm};
  \node[box=blue, minimum width=3.4cm] (F) at (0,\yF) {Feed-forward network};
  \node[box=gray, minimum width=3.4cm] (N2) at (0,\yNb) {Add \& LayerNorm};
  \node[font=\fontsize{8.5}{10}\selectfont\bfseries] (Y) at (0,\yOut) {Output ($N\times d$)};
  \draw[flow] (G) -- node[lbl, right] {$\mathbf{O}^{*}$} (N1);
  \draw[flow] (N1) -- (F); \draw[flow] (F) -- (N2); \draw[flow] (N2) -- (Y);
  \draw[res] (X.east) -- (6.55,\yX) |- node[lbl, pos=0.22, right] {residual} (N1.east);
  \draw[res] (N1.west) -- ++(-1.1,0) |- (N2.west);

  % ---------- MAHA layer frame ----------
  \begin{scope}[on background layer]
    \node[draw=black!40, densely dashed, rounded corners=4pt, fill=black!2,
          fit=(C)(V)(S1)(SL)(lefttap)(lefttap2)(A1)(AL)(U1)(UL)(G), inner sep=7pt] (frame) {};
  \end{scope}
  \node[lbl, anchor=south west, font=\fontsize{7.5}{9}\selectfont\itshape] at ([xshift=2pt,yshift=1pt]frame.north west)
      {MAHA layer (replaces MHSA)};
\end{tikzpicture}
}
    \caption{\rev{Transformer block with MAHA. Attention is computed only at the compressed scales $\mathbf{X}_1,\ldots,\mathbf{X}_L$ (lengths $N/r^l$), obtained sequentially by the downsampling operators $\mathcal{D}_l$. Queries and keys are scale-specific, whereas values are projected once and downsampled with the same operators. The scale outputs are aligned to length $N$ by nearest-neighbor upsampling and combined by simplex-constrained mixing or iterative consensus. Dashed arrows denote residual connections.}}
    \label{fig:architecture}
\end{figure}

The sequence lengths follow an exponential decay schedule $n_l = \lfloor n_{l-1} / r \rfloor$, where $r > 1$ is a compression ratio hyperparameter. This creates a pyramidal structure where higher scales capture increasingly coarse-grained semantic patterns while preserving essential features.

\subsection{Multiscale Attention Computation}
At each scale $l$, MAHA computes independent attention matrices. A key innovation in MAHA is the decoupling of projection parameters to enhance efficiency: while Query ($\mathbf{Q}$) and Key ($\mathbf{K}$) projections are scale-specific, the Value ($\mathbf{V}$) projection is shared across scales. Given the representation $\mathbf{X}_l$, the projections are defined as:

\begin{equation}
\mathbf{Q}_l = \mathbf{X}_l \mathbf{W}_l^Q, \quad \mathbf{K}_l = \mathbf{X}_l \mathbf{W}_l^K
\end{equation}

where $\mathbf{W}_l^Q, \mathbf{W}_l^K \in \mathbb{R}^{d \times d_k}$. The attention weights $\mathbf{A}_l$ are computed via the scaled dot-product:
\[ \mathbf{A}_l = \text{softmax}\left(\frac{\mathbf{Q}_l \mathbf{K}_l^T}{\sqrt{d_k}}\right) \]

Unlike standard transformers, MAHA employs a shared value projection: $\mathbf{V}_{base} = \mathbf{X} \mathbf{W}^V$. The value matrix for scale $l$, denoted as $\mathbf{V}_l$, is obtained by applying the corresponding downsampling operator to the base values: $\mathbf{V}_l = \mathcal{D}_l(\mathbf{V}_{base})$. The scale-specific output $\mathbf{O}_l$ is then:

\[ \mathbf{O}_l = \mathbf{A}_l \mathbf{V}_l \]

This design \rev{reduces the number of value-projection parameters from $L$ matrices to one} while ensuring that the information flow remains consistent across granularity levels.

\subsection{Aggregation of Multiscale Attention Outputs}
{
Let $\mathbf{Z}_l=\mathcal{U}_l(\mathbf{O}_l)\in\mathbb{R}^{n\times d}$ denote the output of scale $l$ after nearest-neighbor upsampling to the original sequence length. For any $\mathbf{w}\in\Delta_L$, define
\begin{equation}
\mathbf{O}(\mathbf{w})=\sum_{l=1}^{L}w_l\mathbf{Z}_l.
\label{eq:aggregate-definition}
\end{equation}
The symbol $\mathbf{O}^*$ denotes the representation returned by the selected aggregation rule; it is not an externally supplied reconstruction target.

\textbf{Simplex-constrained aggregation.} Let $\boldsymbol{\alpha}\in\mathbb{R}^{L}$ be trainable logits. The weights and returned representation are
\begin{equation}
w_l=\frac{\exp(\alpha_l)}{\sum_{j=1}^{L}\exp(\alpha_j)},
\qquad
\mathbf{O}^*=\mathbf{O}(\mathbf{w}).
\label{eq:simplex-aggregation}
\end{equation}
Softmax enforces $\mathbf{w}\in\Delta_L$. Hence, $\mathbf{O}^*$ is a convex combination of the aligned outputs, but the logits are learned through the downstream loss rather than by solving a separate convex program. Moreover, $\|\mathbf{w}\|_1=1$ on $\Delta_L$, so the implemented $\ell_1$ term is constant and cannot induce sparsity.

\textbf{Iterative-consensus aggregation.} This input-dependent variant initializes $w_l^{(0)}=1/L$. At iteration $k$, it computes
\begin{align}
\mathbf{O}^{(k)} &= \sum_{l=1}^{L}w_l^{(k)}\mathbf{Z}_l, \\
e_l^{(k)} &= \|\mathbf{Z}_l-\mathbf{O}^{(k)}\|_F, \\
q_l^{(k)} &= -e_l^{(k)}, \\
w_l^{(k+1)} &= \frac{\exp(q_l^{(k)}/\tau)}{\sum_{j=1}^{L}\exp(q_j^{(k)}/\tau)}.
\label{eq:consensus-update}
\end{align}
Here, $\tau>0$ is the response temperature. The implementation uses $K=3$ unrolled updates and $\tau=1$, then returns $\mathbf{O}^*=\mathbf{O}(\mathbf{w}^{(K)})$. No tolerance-based convergence test is used; therefore, the terminal iterate is not asserted to be a fixed point or Nash equilibrium.

\textbf{Relationship to training.} Let $\Theta$ contain the backbone parameters and, for simplex-constrained aggregation, $\boldsymbol{\alpha}$. The released training code computes
\begin{equation}
\begin{aligned}
\min_{\Theta}\quad &\mathcal{L}_{\mathrm{task}}
\!\left(h_{\Theta}(\mathbf{O}^*(\mathbf{x};\Theta)),y\right)
+\gamma\mathcal{L}_{\mathrm{aux}},\\
\mathcal{L}_{\mathrm{aux}}&=
\begin{cases}
\displaystyle\sum_{m=1}^{M}\lambda_s\|\mathbf{w}_m\|_1=M\lambda_s, & \text{simplex constrained},\\
0, & \text{iterative consensus},
\end{cases}
\end{aligned}
\label{eq:training-objective}
\end{equation}
where $h_{\Theta}$ is the task head and $M$ is the number of MAHA layers. The released scripts use $\gamma=0.1$ and $\lambda_s=0.1$. Because the simplex-constrained auxiliary term is constant, it has zero gradient and does not provide sparsity regularization. Task-loss gradients pass through Eq.~\eqref{eq:simplex-aggregation} or through the $K$ unrolled updates in Eq.~\eqref{eq:consensus-update}.
}

\subsection{Hybrid Dilated-Convolutional Transformer Design}
MAHA applies a dilated convolution with a ReLU activation to the input before hierarchical decomposition, thereby enriching local context without introducing a separate cross-scale gating stage. The resulting sequence is decomposed into the $L$ resolutions described in Section~4.1, processed by scale-specific attention, aligned to the original sequence length through nearest-neighbor upsampling, and combined by one of the aggregation rules in Section~4.3. This description matches the released reference implementation; the earlier reference to a cross-scale gate was removed because that module is not part of the executed MAHA block.

\subsection{Complexity Reduction through Hierarchical Compression}
The total computational complexity of MAHA is governed by the hierarchical decomposition. Since attention is evaluated on the compressed scales $l=1,\ldots,L$ rather than on a full $n\times n$ matrix, the cost for a sequence of length $n$ is:

\begin{equation}
T(n) = \sum_{l=1}^L \left( \frac{n}{r^l} \right)^2 d + O(n \log n)
\end{equation}

Evaluating the finite geometric series yields:

\begin{equation}
T(n) = n^2 d\,\frac{1-r^{-2L}}{r^2-1} + O(n\log n)
\end{equation}

For fixed $r$ and $L$, the leading term remains quadratic, $O(n^2/(r^2-1))$, but its coefficient is lower than that of full-resolution attention. For example, as $L$ grows and $r=2$, the attention term approaches $n^2d/3$ rather than $n^2d$. \rev{For the configuration used in this study ($r=2$, $L=4$, scales $n/2,\ldots,n/16$), the coefficient is $(1-2^{-8})/3\approx0.332$; that is, MAHA computes 66.8\% fewer attention scores than full-resolution attention, independently of $n$.} Accordingly, MAHA should be interpreted as a coefficient-reduced quadratic method under the stated hierarchy, not as an asymptotically linear attention mechanism.

\begin{itemize}
    \item \textbf{Scale-Specific Sparsity:} Coarser scales have $n_l \ll n$, reducing the cost quadratically.
    \item \textbf{Normalized Scale Mixing:} Both aggregation rules produce nonnegative weights that sum to one; this constraint does not itself imply sparse scale selection.
\end{itemize}

\begin{figure}[htbp]
    \centering
    \begin{tikzpicture}[x=0.72cm,y=0.62cm]
        \draw[->,thick] (0,0) -- (8.2,0) node[below] {Sequence length $N$};
        \draw[->,thick] (0,0) -- (0,6.2) node[above,align=center] {Relative\\attention cost};
        \draw[red,very thick,domain=0:7.3,samples=80]
            plot (\x,{0.105*\x*\x});
        \draw[green!55!black,very thick,domain=0:7.3,samples=80]
            plot (\x,{0.035*\x*\x});
        \node[red,anchor=west] at (5.15,4.85) {Standard attention: $cN^2$};
        \node[green!45!black,anchor=west,align=left] at (5.15,1.35)
            {MAHA: $c' N^2$\\with $c'<c$};
        \draw[densely dashed,gray] (7.3,0) -- (7.3,5.6);
        \draw[<->,thick] (7.0,1.86) -- (7.0,5.15)
            node[midway,left,align=right] {coefficient\\reduction};
    \end{tikzpicture}
    \caption{Computational cost comparison over the evaluated sequence-length range. Hierarchical compression reduces MAHA's quadratic coefficient relative to standard self-attention; it does not change the asymptotic order under a fixed compression ratio.}
    \label{fig:complexity}
\end{figure}

\section{Experiments}

The experiments were organized around three questions. First, we examined whether MAHA could improve computational efficiency while retaining competitive performance on downstream tasks (RQ1). Second, we compared simplex-constrained and iterative-consensus aggregation to determine how the choice of aggregation rule affects model behavior (RQ2). Third, we investigated how the number and granularity of hierarchical scales influence the balance between representational accuracy and computational cost (RQ3).

\subsection{Experimental Setup}
We evaluated MAHA on four established benchmarks representing complementary sequence-modeling requirements. MNLI and SST-2 from the GLUE benchmark were used for text classification \citep{wang2018a}, whereas PG-19 provided a long-context language-modeling setting with sequences exceeding 4,000 tokens \citep{rae2020a}. WMT14 English--German was selected for machine translation \citep{bojar2014a}, and SQuAD v2.0 was used to assess question answering \citep{rajpurkar2018a}. Together, these datasets cover sentence-level understanding, long-range dependency modeling, sequence generation, and extractive reasoning.

The comparison included standard Multi-Head Attention (MHA) \citep{vaswani2017a} and four efficient-attention baselines: Longformer \citep{beltagy2020a}, BigBird \citep{zaheer2020a}, Reformer \citep{kitaev2020a}, and Performer \citep{choromanski2020a}. The experimental backbone comprised 12 Transformer layers with a hidden dimensionality of 768 and 12 attention heads. We used a batch size of 32 for classification and 16 for translation and question answering, with a learning rate of $5 \times 10^{-5}$ and 10,000 warm-up steps. Sequence length was set to 512 for classification and question answering and to 4096 for PG-19. \rev{For MAHA, we used $L=4$ compressed scales of length $n/2$, $n/4$, $n/8$, and $n/16$ (256, 128, 64, and 32 tokens for $n=512$; 2048, 1024, 512, and 256 tokens for $n=4096$)} and strided convolution with a kernel size of 3. The iterative-consensus variant used $K=3$ updates and $\tau=1$. The released scripts retain $\gamma=0.1$ and $\lambda_s=0.1$ in Eq.~\eqref{eq:training-objective}; because these coefficients multiply a constant in the simplex-constrained branch, they do not induce sparsity.

\rev{All six attention mechanisms in Table~\ref{tab:performance} were trained by the author in the same 12-layer backbone with the same data, sequence lengths, and optimization settings; only the attention layer differs. Each configuration in Table~\ref{tab:performance}, Table~\ref{tab:ablation}, and Figure~\ref{fig:scales} was trained once with a single fixed random seed. Repeated runs were not feasible within the available compute budget; the reported values are therefore single-run point estimates, and no standard deviations are available.}

\subsection{\rev{Comparability of Reported Results}}
\rev{Table~\ref{tab:comparability} lists every source of quantitative evidence in this paper together with its data, sequence length, model size, and protocol. Only results produced under the common protocol of Section~5.1 are compared numerically. Results published for the recent methods in Table~\ref{tab:related-families} were obtained with different corpora, context lengths (often tens of thousands of tokens or more), model sizes, tokenizers, training budgets, and hardware. Placing those numbers next to Table~\ref{tab:performance} would confound architectural differences with differences in scale and protocol; they are therefore discussed qualitatively in Section~2.2 and not reproduced as scores.}

\begin{table*}[htbp]
\centering
\small
\caption{\rev{Normalized overview of the quantitative evidence reported in this paper. ``Common protocol'' denotes the backbone and training settings of Section~5.1 (12 layers, $d=768$, 12 heads, learning rate $5\times10^{-5}$, 10,000 warm-up steps, single seed).}}
\label{tab:comparability}
{\color{revisionblue}
\setlength{\tabcolsep}{3pt}\begin{tabular}{>{\raggedright\arraybackslash}p{0.18\linewidth}>{\raggedright\arraybackslash}p{0.14\linewidth}>{\raggedright\arraybackslash}p{0.15\linewidth}>{\raggedright\arraybackslash}p{0.11\linewidth}>{\raggedright\arraybackslash}p{0.14\linewidth}>{\raggedright\arraybackslash}p{0.13\linewidth}}
\toprule
\textbf{Evidence} & \textbf{Data / task} & \textbf{Seq.\ length} & \textbf{Model size} & \textbf{Protocol} & \textbf{Directly comparable?} \\
\midrule
Table~\ref{tab:performance}: MHA, Longformer, BigBird, Reformer, Performer, MAHA & MNLI, SST-2, PG-19, WMT14 En--De, SQuAD v2.0 & Task-specific, identical across models (512 for MNLI, SST-2, SQuAD; 4096 for PG-19) & Identical backbone for all models & Common protocol, trained by the author, single run & Yes, within each column \\
Table~\ref{tab:ablation} and Figure~\ref{fig:scales}: MAHA variants & MNLI & 512 & Common backbone & Common protocol, single run & Yes, relative to default MAHA \\
Figure~\ref{fig:efficiency}: attention cost & Analytical count & 1K--8K & Per head, independent of $d$ & Query--key scores only; $O(n)$ overhead excluded & Not a runtime measurement \\
Figure~\ref{fig:ablation}: loss curves & Synthetic random tokens & 64 & 1 layer, $d=64$, $L=3$ & Released demonstration script, 5 epochs & Illustrative only \\
Recent methods (Table~\ref{tab:related-families}) & Corpora of the original papers & Varies by paper & Varies by paper & Original authors' settings and hardware & No; discussed qualitatively \\
\bottomrule
\end{tabular}
}
\end{table*}

\subsection{Main Results}
Table \ref{tab:performance} summarizes the performance on benchmark datasets. \rev{MAHA remains within 0.2 points of standard MHA on MNLI, SST-2, WMT, and SQuAD and obtains the lowest PG-19 perplexity among the compared models. Because every entry is a single run, differences of a few tenths of a point, including the 0.4-point perplexity margin over BigBird on PG-19, are indicative rather than statistically established.}

\begin{table*}[htbp]
    \centering
    \setlength{\tabcolsep}{5.8pt}
    \caption{Performance Comparison Across Tasks \rev{(single run per model; common backbone and protocol of Section~5.1)}.}
    \label{tab:performance}
    \begin{tabular}{lccccc}
        \toprule
        \textbf{Model} & \textbf{MNLI} & \textbf{SST-2} & \textbf{PG-19} & \textbf{WMT} & \textbf{SQuAD} \\
         & (Acc) & (Acc) & (PPL) $\downarrow$ & (BLEU) & (F1) \\
        \midrule
        Standard MHA & 86.2 & 93.5 & 24.3 & 28.7 & 88.4 \\
        Longformer & 85.7 & 92.8 & 23.8 & 27.9 & 87.6 \\
        BigBird & 85.9 & 93.1 & 23.5 & 28.1 & 87.9 \\
        Reformer & 84.3 & 91.7 & 25.6 & 26.4 & 85.2 \\
        Performer & 85.1 & 92.4 & 24.9 & 27.3 & 86.7 \\
        \textbf{MAHA (Ours)} & \textbf{86.0} & \textbf{93.3} & \textbf{23.1} & \textbf{28.5} & \textbf{88.2} \\
        \bottomrule
    \end{tabular}
    \small
    \vspace{0.1cm} \\
    \textit{\rev{Note: Higher is better except for PG-19 perplexity. Values are single-run point estimates without standard deviations.}}
\end{table*}

\subsection{Computational Efficiency Analysis}
\rev{Efficiency is assessed analytically as the number of attention-score (query--key) computations per head, which is $N^2$ for full-resolution attention and $\sum_{l=1}^{4}(N/2^l)^2\approx0.332N^2$ for the MAHA configuration of Section~5.1. Wall-clock throughput and peak memory were not measured under a common procedure for all baselines and are therefore not reported. As illustrated in Figure~\ref{fig:efficiency}, hierarchical compression lowers the cost by a constant factor over the whole range, while both curves remain quadratic.}

\begin{figure}[htbp]
    \centering
    \begin{tikzpicture}[x=0.72cm,y=0.070cm]
        \draw[->,thick] (0.7,0) -- (8.6,0);
        \node[below] at (4.7,-8) {Sequence length ($\times 1024$)};
        \draw[->,thick] (0.8,0) -- (0.8,78) node[above,align=center] {Attention scores per head\\(millions)};
        \foreach \x in {1,...,8} {
            \draw[gray!35] (\x,0) -- (\x,72);
            \node[below] at (\x,0) {\scriptsize \x K};
        }
        \foreach \y in {10,20,...,70} {
            \draw[gray!35] (0.8,\y) -- (8.35,\y);
            \node[left] at (0.8,\y) {\scriptsize \y};
        }
        \draw[red,very thick] plot[smooth] coordinates
            {(1,1.05) (2,4.19) (3,9.44) (4,16.78) (5,26.21) (6,37.75) (7,51.38) (8,67.11)};
        \draw[green!55!black,very thick] plot[smooth] coordinates
            {(1,0.35) (2,1.39) (3,3.13) (4,5.57) (5,8.70) (6,12.53) (7,17.06) (8,22.28)};
        \draw[densely dashed] (4,0) -- (4,16.8);
        \fill[red] (4,16.8) circle (2pt) node[above left] {\scriptsize 16.8M};
        \fill[green!55!black] (4,5.57) circle (2pt) node[below right] {\scriptsize 5.57M};
        \draw[<->,thick] (3.85,5.57) -- (3.85,16.8);
        \node[anchor=west] at (1.1,42) {\scriptsize MAHA: 66.8\% fewer scores};
        \fill[white] (1.0,59) rectangle (4.6,73);
        \draw[red,very thick] (1.05,69) -- (1.55,69)
            node[right,black] {\scriptsize Standard MHA ($N^2$)};
        \draw[green!55!black,very thick] (1.05,63) -- (1.55,63)
            node[right,black] {\scriptsize MAHA ($0.332N^2$)};
    \end{tikzpicture}
    \caption{\rev{Analytical attention-score computations per head versus sequence length. MAHA ($L=4$, $r=2$) evaluates $0.332N^2$ query--key scores, 66.8\% fewer than full attention at every $N$ (5.57M versus 16.8M at $N=4096$). Both curves are quadratic; downsampling and aggregation overhead is not included.}}
    \label{fig:efficiency}
\end{figure}

\rev{At $N=4096$, full attention computes $4096^2\approx16.8$M query--key scores per head, whereas MAHA computes $2048^2+1024^2+512^2+256^2\approx5.57$M, i.e., 66.8\% fewer. These are analytical operation counts rather than measured runtimes: the dilated convolution, downsampling, upsampling, and aggregation add $O(n)$ terms that are not included, and the constant-factor reduction does not change the asymptotic order established in Section~4.5.}

\subsection{Ablation Studies}
\rev{We evaluated the contribution of the aggregation rule and of the hierarchy depth on MNLI. Table~\ref{tab:ablation} collects all component-level results relative to the default configuration; every entry is a single run.}

\subsubsection{Aggregation Strategy Comparison}\label{sec:aggregation-ablation}
\begin{figure}[htbp]
    \centering
     \includegraphics[width=0.9\linewidth]{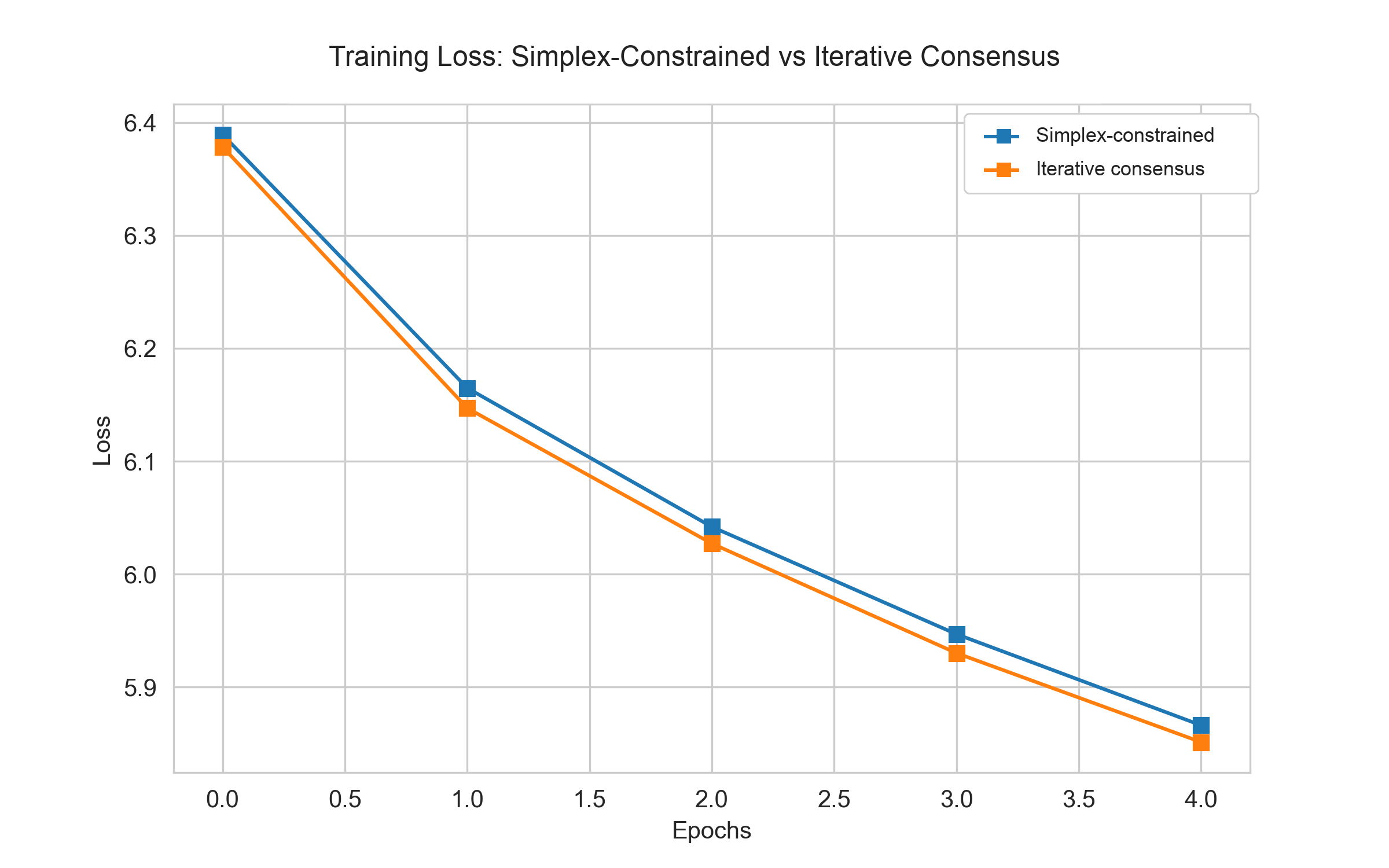}
     \caption{\rev{Training-loss comparison between simplex-constrained mixing and finite-step iterative consensus on synthetic random-token data (released demonstration script; illustrative only).}}
     \label{fig:ablation}
\end{figure}

\rev{Figure~\ref{fig:ablation} shows training-loss curves for both aggregation rules produced by the lightweight demonstration script of the released repository, which trains a one-layer model ($d=64$, $L=3$) on randomly generated token sequences for five epochs. The curves therefore illustrate the optimization behavior of the two rules and are not MNLI learning curves. In this demonstration, iterative consensus (orange) reaches a marginally lower loss than simplex-constrained mixing (blue); this trajectory is not evidence of convergence to a game-theoretic equilibrium.}

\rev{On MNLI, simplex-constrained mixing obtained the highest accuracy (86.0\%), followed by iterative consensus (85.8\%) and uniform mean aggregation with fixed weights $w_l=1/L$ (85.2\%) (Table~\ref{tab:ablation}). Learned scale weighting thus contributed 0.8 points over uniform averaging in this single-run comparison. Iterative consensus adds $K$ unrolled updates over the $L$ aligned outputs, i.e., $O(KLnd)$ additional operations per layer relative to direct mixing.}

\begin{table}[htbp]
    \centering
    \caption{\rev{Component-level ablation of MAHA on MNLI (single run per configuration). $\Delta$ is the change relative to the default configuration.}}
    \label{tab:ablation}
    {\color{revisionblue}
    \begin{tabular}{llcc}
        \toprule
        \textbf{Component} & \textbf{Variant} & \textbf{MNLI (Acc)} & \textbf{$\Delta$} \\
        \midrule
        Default & Simplex mixing, $L=4$ & 86.0 & -- \\
        \midrule
        Aggregation rule & Iterative consensus ($K=3$, $\tau=1$) & 85.8 & $-0.2$ \\
         & Mean aggregation ($w_l=1/L$) & 85.2 & $-0.8$ \\
        \midrule
        Hierarchy depth & $L=2$ & 84.5 & $-1.5$ \\
         & $L=3$ & 85.4 & $-0.6$ \\
         & $L=5$ & 85.3 & $-0.7$ \\
         & $L=6$ & 84.8 & $-1.2$ \\
        \bottomrule
    \end{tabular}
    }
\end{table}

\subsubsection{Scale Configuration Analysis}
\begin{figure}[htbp]
    \centering
    \includegraphics[width=0.7\linewidth]{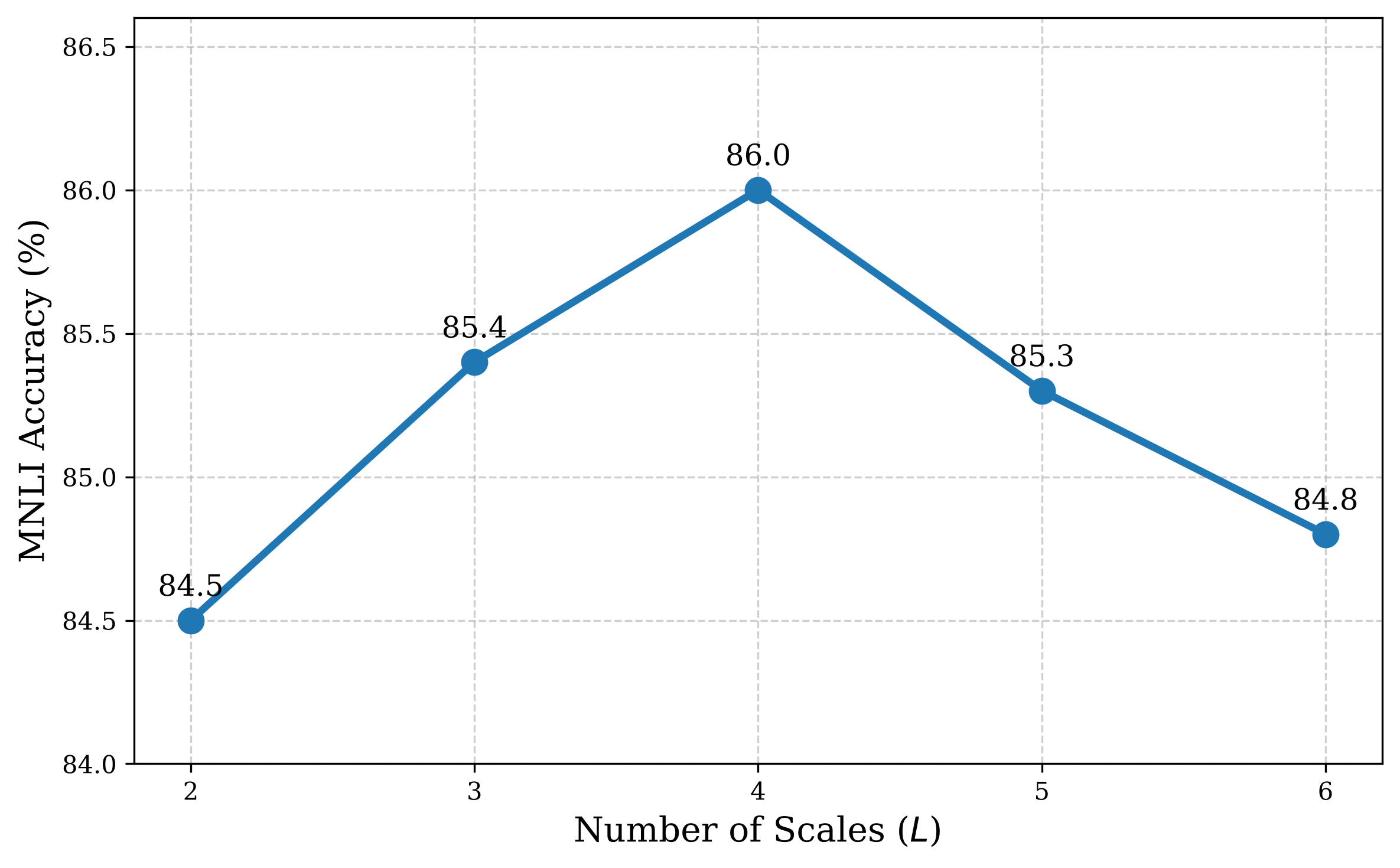} 
    \caption{MNLI Accuracy vs Number of Hierarchical Scales ($L$). Optimal performance is at $L=4$ \rev{(single run per configuration)}.}
    \label{fig:scales}
\end{figure}

We analyzed how the depth of the hierarchy (number of scales, $L$) affects model performance. As illustrated in Figure \ref{fig:scales}, optimal results are observed at $L=4$, balancing granularity and context. \rev{Using too few scales ($L=2$) reduced accuracy to 84.5\%, and deeper hierarchies also reduced it ($L=5$: 85.3\%; $L=6$: 84.8\%), possibly because the coarsest scales become very short (16 and 8 tokens for $n=512$).}

\rev{\textbf{Components not isolated.} The dilated convolution, the shared value projection, and the choice between strided-convolution and adaptive-pooling downsampling were not ablated individually in this study. Their individual contributions therefore remain unquantified; Section~6.2 lists this as a limitation.}

\subsection{\rev{Illustrative Scale Patterns}}
\begin{figure}[htbp]
    \centering
    \includegraphics[width=0.8\linewidth]{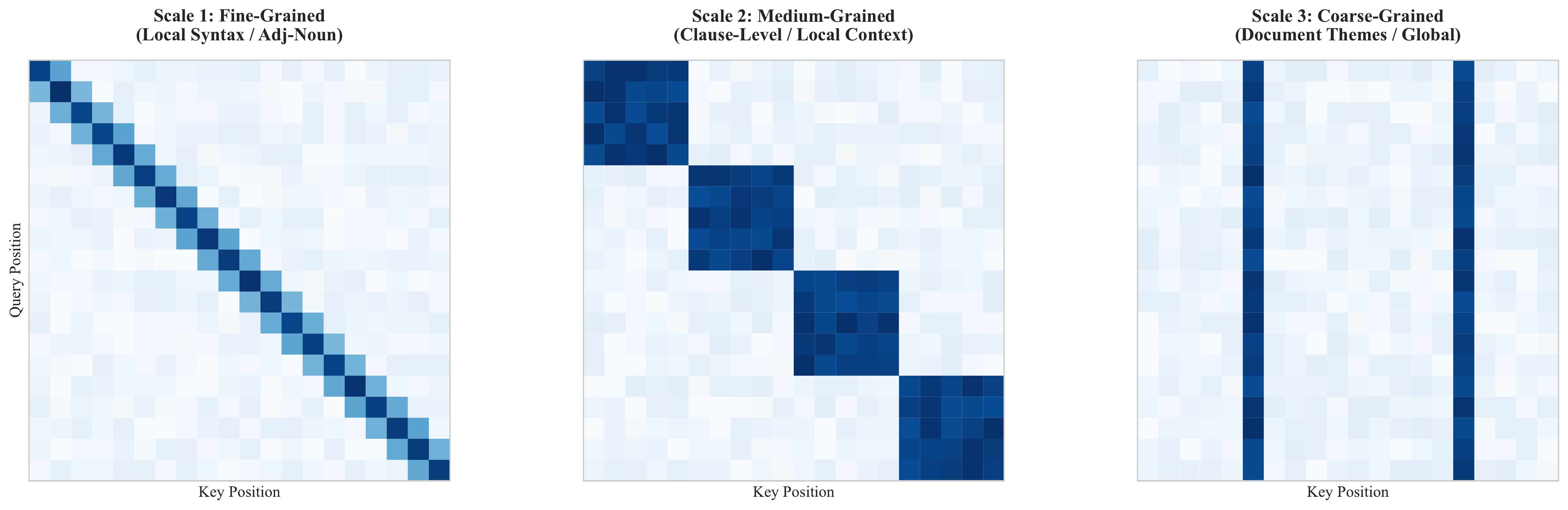} 
    \caption{\rev{Schematic (synthetic) illustration of the intended fine-, medium-, and coarse-scale attention patterns. The matrices are constructed for exposition and are not learned attention weights.} Darker regions indicate higher attention weights.}
    \label{fig:heatmap}
\end{figure}

\rev{Figure~\ref{fig:heatmap} is a schematic illustration of the attention patterns that the hierarchy is designed to support. The matrices were constructed synthetically for exposition and are not attention weights extracted from a trained model.}

\rev{At the finest scale, mass is concentrated near the diagonal, corresponding to local relations such as adjective--noun pairs. At an intermediate scale, a block-diagonal pattern corresponds to clause-level interactions. At the coarsest scale, vertical bands correspond to a few positions attended from across the sequence. Whether trained MAHA layers exhibit these patterns requires extracting learned weights and is left for future work.}

\section{Discussion}

\subsection{Scalability vs. Implementation Overhead}
Our experiments highlight a critical distinction between asymptotic order, its leading coefficient, and implementation overhead. Under a fixed compression ratio and number of scales, both MAHA and standard attention retain an $O(N^2)$ leading term; MAHA reduces its coefficient by evaluating attention only at compressed resolutions. Figure \ref{fig:efficiency} therefore \rev{quantifies an analytical constant-factor reduction} rather than evidence of $O(N)$ asymptotic growth. \rev{Realized speed and memory gains also depend on the implementation, aggregation rule, and downsampling overhead, and they were not measured under a common procedure in this study.}

\subsection{Limitations}
MAHA's practical use is subject to several constraints. The number of hierarchical scales $L$ and the compression ratio $r$ must be selected for the target domain, and inappropriate settings may either discard fine-grained information or add unnecessary computation. The finite-step iterative-consensus rule introduces computational overhead relative to direct simplex-constrained mixing. Its terminal iterate is not guaranteed to be a fixed point, and the present study does not provide a convergence theorem. More fundamentally, MAHA assumes that relevant linguistic structure can be represented across a hierarchy of resolutions. This assumption may be less suitable for non-compositional expressions or dispersed references in highly unstructured text.

\rev{The empirical evidence has further limitations. All results in Tables~\ref{tab:performance} and~\ref{tab:ablation} and in Figure~\ref{fig:scales} are single-seed point estimates; seed-to-seed variation was not estimated, so differences of a few tenths of a point should not be interpreted as reliable. The common-protocol comparison covers four efficient-attention baselines published up to 2020, whereas the recent methods discussed in Section~2.2, such as NSA, MoBA, GLA, and Mamba, were not retrained under the same protocol. In addition, the individual contributions of the dilated convolution, the shared value projection, and the downsampling operator were not isolated. Multi-seed evaluation reporting mean $\pm$ standard deviation, a common-protocol comparison with recent long-context methods, and a complete component ablation are priorities for future work.}

\subsection{Potential Application Scenarios}
The multiscale structure of MAHA may also be useful outside conventional NLP, although such extensions require task-specific validation. In genomics, modeling dependencies across multiple resolutions could support the analysis of long DNA sequences, where \rev{relevant interactions can be long-range} \citep{choi2023a}. In multimodal learning, \rev{hierarchical transformers have been applied to video--text retrieval \citep{liu2021a}, and the multiscale hierarchy could similarly be adapted to temporal video resolutions such as frames, shots, and scenes.} The normalized aggregation may be relevant to federated-learning settings in which clients contribute information at different granularities or levels of quality \citep{chen2020a}; however, this possibility remains to be evaluated empirically.

\subsection{Ethical Considerations}
The potential efficiency gains of MAHA present a dual-edged sword. Lower computational demand may reduce energy use and associated emissions per run, although environmental impact was not measured in this study \citep{strubell2019a}; lower costs may also incentivize the training of larger or additional models. Furthermore, hierarchical aggregation introduces interpretability challenges: although scale-specific weights can be inspected, their interaction may obscure the model's decision-making path \citep{danilevsky2021a}. Future work should address these transparency issues and quantify energy consumption before making environmental claims.

\section{Conclusion}

In this paper, we introduced Multiscale Aggregated Hierarchical Attention (MAHA), a framework that decomposes sequences into hierarchical granularities and combines the aligned outputs through simplex-constrained mixing or finite-step iterative consensus. The revised formulation defines the returned representation, update equations, hyperparameters, and training objective without claiming global convexity or a computed Nash equilibrium. In the reported evaluation, MAHA reduces the leading coefficient of quadratic attention \rev{(66.8\% fewer attention-score computations for the configuration studied)} while retaining competitive contextual modeling performance.

In the reported evaluation, MAHA obtained the lowest PG-19 perplexity among the compared methods and competitive results on classification, machine translation, and question answering. \rev{Because these are single-run results, they should be confirmed with multiple seeds and a broader set of recent baselines.} The proposed hybrid dilated-convolutional architecture is designed as a replacement for existing attention layers.

Looking forward, \rev{MAHA may contribute to more efficient foundation models in resource-constrained environments.} Future work should examine convergence conditions for the iterative update and evaluate regularizers that can genuinely promote selective scale use. The framework may also be extended to modalities such as computer vision and speech processing, where multiscale representation is relevant.

\section*{Statements and Declarations}

\textbf{Funding} \\
The author declares that no funds, grants, or other support were received during the preparation of this manuscript.

\bigskip

\textbf{Competing Interests} \\
The author has no relevant financial or non-financial interests to disclose.

\bigskip

\textbf{Author Contributions} \\
The author confirms sole responsibility for the following: study conception and design, development of the mathematical aggregation framework, implementation of the MAHA architecture, data analysis, and manuscript preparation.

\bigskip

\textbf{Data Availability} \\
The publicly available datasets analysed in this study are cited in the manuscript. A reference implementation of the MAHA layers, together with configuration files and lightweight synthetic demonstration scripts, is available in the GitHub repository: \url{https://github.com/canererden/MAHA-Project}. The repository does not contain end-to-end data-preparation and training pipelines for independently regenerating every benchmark value reported in Table~\ref{tab:performance}.

\bibliography{references}

\end{document}